\documentclass[letterpaper, 10 pt, conference, twoside]{ieeeconf} 
\usepackage{graphicx}
\usepackage{amsmath}
\usepackage{bbding}
\usepackage{pifont}
\usepackage{wasysym}
\usepackage{amssymb}
\usepackage{hyperref}
\usepackage{threeparttable} 
\usepackage{cite} 
\usepackage{booktabs}
\usepackage{xcolor}
\usepackage{multirow}
\usepackage{blindtext}
\usepackage{flushend} 
\hypersetup{hidelinks}   
\usepackage{bm}
\usepackage{subcaption}
\usepackage[linesnumbered, ruled]{algorithm2e}

\SetKwInput{KwInput}{Input}
\SetKwInput{KwOutput}{Output}

\usepackage{cite}

\IEEEoverridecommandlockouts                              

\SetKwComment{Comment}{$\triangleright$\ }{}
\usepackage{afterpage}
\usepackage{mwe}

\title{\LARGE \bf CenterLineDet: CenterLine Graph Detection for Road Lanes with Vehicle-mounted Sensors by Transformer for HD Map Generation}

\author{Zhenhua Xu\textsuperscript{1}, Yuxuan Liu\textsuperscript{1}, Yuxiang Sun\textsuperscript{2}, Ming Liu\textsuperscript{1,3,4}, and Lujia Wang\textsuperscript{1} %
\thanks{\textsuperscript{1}The Hong Kong University of Science and Technology, Clear Water Bay, Kowloon, Hong Kong (email: \{zxubg,yliuhb\}@connect.ust.hk; \{eelium,eewang\}@ust.hk).}
\thanks{\textsuperscript{2}The Hong Kong Polytechnic University, Hung Hom, Kowloon, Hong Kong (e-mail: yx.sun@polyu.edu.hk, sun.yuxiang@outlook.com).}
\thanks{\textsuperscript{3}The Hong Kong University of Science and Technology (Guangzhou), Nansha, Guangzhou, China (e-mail: eelium@ust.hk).}
\thanks{\textsuperscript{4}HKUST Shenzhen-Hong Kong Collaborative Innovation Research Institute, Futian, Shenzhen, China (e-mail: eelium@ust.hk).}
\thanks{{Corresponding author: Lujia Wang.}}
} 

\begin{document}
\maketitle


\begin{abstract}
With the fast development of autonomous driving technologies, there is an increasing demand for high-definition (HD) maps, which provide reliable and robust prior information about the static part of the traffic environments. As one of the important elements in HD maps, road lane centerline is critical for downstream tasks, such as prediction and planning. Manually annotating centerlines for road lanes in HD maps is labor-intensive, expensive and inefficient, severely restricting the wide applications of autonomous driving systems. Previous work seldom explores the lane centerline detection problem due to the complicated topology and severe overlapping issues of lane centerlines. In this paper, we propose a novel method named CenterLineDet to detect lane centerlines for automatic HD map generation. Our CenterLineDet is trained by imitation learning and can effectively detect the graph of centerlines with vehicle-mounted sensors (i.e., six cameras and one LiDAR) through iterations. Due to the use of the DETR-like transformer network, CenterLineDet can handle complicated graph topology, such as lane intersections. The proposed approach is evaluated on the large-scale public dataset NuScenes. The superiority of our CenterLineDet is demonstrated by the comparative results. 
Our code, supplementary materials, and video demonstrations are available at \href{https://tonyxuqaq.github.io/projects/CenterLineDet/}{https://tonyxuqaq.github.io/projects/CenterLineDet/}.
    
\end{abstract}


\section{Introduction}
	High-definition (HD) maps are critical to autonomous driving vehicles since they provide reliable information about the static part of traffic environments. HD maps have many layers that consist of various line-shaped road elements. Lower-level layers consist of physically existing elements (e.g., road boundaries, road curbs, and road lanelines), while high-level layers have virtual elements (e.g., road lane centerlines). All of the aforementioned HD map layers are recorded in vector data format (i.e., graphs with vertices and edges). Road elements in low-level layers of the HD map are usually utilized to prevent potential collisions and assure the safety of vehicles. High-level layers of HD maps can define the path that vehicles can drive on. They contain all the topology information of roads, thus they are important for downstream tasks, such as vehicle planning, prediction and control \cite{gao2020vectornet,liang2020learning,da2022path}.  At this stage, creating the HD map of a target region heavily relies on human annotators, which is labor-intensive, inefficient, and expensive. Therefore, an approach that can automatically create the HD map with road lane centerlines is of great interest to the research communities. Unlike low-level road elements, road lane centerlines have complicated topology structures (e.g., intersections) and severe overlapping issues, thus making the detection of centerline graphs challenging.
	
	To the best of our knowledge, taken as input multi-frame sequence collected by vehicle-mounted sensors, most previous works only focus on the detection task in a single frame and output rasterized results \cite{li2021hdmapnet,pan2020cross,zhou2022cross,can2021structured,philion2020lift,can2021topology}, which does not meet the requirement for HD map building  that demands global vectorized detection results.
	
	Even though some previous works seek to detect road elements for vectorized map generation purposes, such as road network graph detection \cite{xu2022rngdet,xu2022rngdet++,bastani2018roadtracer}, road laneline graph detection \cite{homayounfar2018hierarchical,homayounfar2019dagmapper} and road boundary detection \cite{xu2021topo,liang2019convolutional,xu2021csboundary,zhxu2021icurb,xu2021cp}, they rely on bird's-eye view (BEV) aerial images captured by satellites or UAVs, instead of data collected by vehicle-mounted sensors that we discuss in this work. 
	
	To provide a solution to the aforementioned problems, in this paper, we present CenterLineDet (i.e., Lane \underline{CenterLine} Graph \underline{Det}ector), a DETR-like \cite{carion2020end} model that detects the global lane centerline graph with vehicle-mounted sensors for multi-frame and long-term HD map generation purpose. Our CenterLineDet first fuses data collected by sensors in multiple frames and projects it to BEV, then iteratively generates the global HD map by a trained DETR-like decision-making transformer network. CenterLineDet works on sequential data and does not require pre-built point cloud maps like some past works \cite{homayounfar2019dagmapper,liang2019convolutional}.
	
 The contributions of this work are listed as follows:
	\begin{itemize} 
	    \item We present CenterLineDet, an effective deep learning approach that automatically creates the global HD map of lane centerlines with sequential data captured by vehicle-mounted sensors as input.
	    \item We evaluate CenterLineDet on the large publicly available dataset NuScenes \cite{caesar2020nuscenes} to demonstrate the superiority of our approach.
	\end{itemize}


\section{Related Works}
\label{sec:related work}
\subsection{Applications of Road Lane Centerline HD map}

Road lane centerlines are virtual lines defined by humans based on road topology, road connectivity, and traffic rules. Thus, lane centerlines contain abundant information about roads, which makes it critical for plenty of downstream tasks of autonomous vehicles, such as motion prediction \cite{gao2020vectornet,liang2020learning,da2022path,liu2021role}, and vehicle navigation (i.e., planning and control) \cite{zhou2021automatic}. 
Christensen \textit{et al.} \cite{christensen2021autonomous} proposed an autonomous driving system for micro-mobility. The global planner and local planner of this system heavily rely on the lane centerline HD map. For the global planner, the centerline HD map was used to calculate the shortest path to the destination since it contained all the topology and connectivity information of the road network. For the local planner and controller, the vehicle was controlled to follow the lane centerline ahead (the centerline HD map defines the path that vehicles can drive on). 
Liang \textit{et al.} \cite{liang2020learning} extracted features of lane centerline HD maps by a graph neural network as prior information to assist the motion prediction of objects on the road. 

\subsection{Road Element Detection with Vehicle-mounted Sensors}

Most previous works resort to end-to-end perspective transformation to detect road elements in BEV \cite{li2021hdmapnet,pan2020cross,zhou2022cross,philion2020lift,can2021structured,can2021topology}. In these works, with data collected by vehicle-mounted sensors as input, a deep learning network is trained to fuse the data and produce the probabilistic distribution of target elements in the BEV. 
Li \textit{et al.} \cite{li2021hdmapnet} fused six vehicle-mounted cameras together with a LiDAR, and trained an end-to-end deep neural network to predict the BEV segmentation map of road lanelines. Based on the segmentation results of the BEV image, the authors vectorized the segmented lanelines by the skeletonization algorithm to obtain the final road laneline graph. 
Can \textit{et al.} \cite{can2021structured,can2021topology} modeled the lane centerline by B-splies, and predicted splines in the current frame by a DETR-like network.

To the best of our knowledge, most aforementioned works only focus on the detection task in a single frame \cite{can2021structured,can2021topology,li2021hdmapnet}, leaving the problem of merging local maps of multiple frames into a global map (i.e., long-term mapping problem) unexplored. Moreover, their task is the detection of simple road elements without complicated topology changes or overlapping issues, such as road boundaries and road lanelines \cite{elhousni2020automatic}. To further improve the detection results of road elements, some works resort to additional data like OpenStreetMap (OSM) \cite{zhou2021automatic,joshi2015generation} for enhancement. However, all the above works cannot well handle the following problems of lane centerline HD map generation: (1) how to handle complicated topology and overlapping issues, especially within the road intersection areas; (2) how to merge detection results of each frame into the final global vector HD map.


\section{Methodology}
\label{sec:methodology}

\subsection{Overview}
\begin{figure*}[t]
  \centering
  \includegraphics[width=\linewidth]{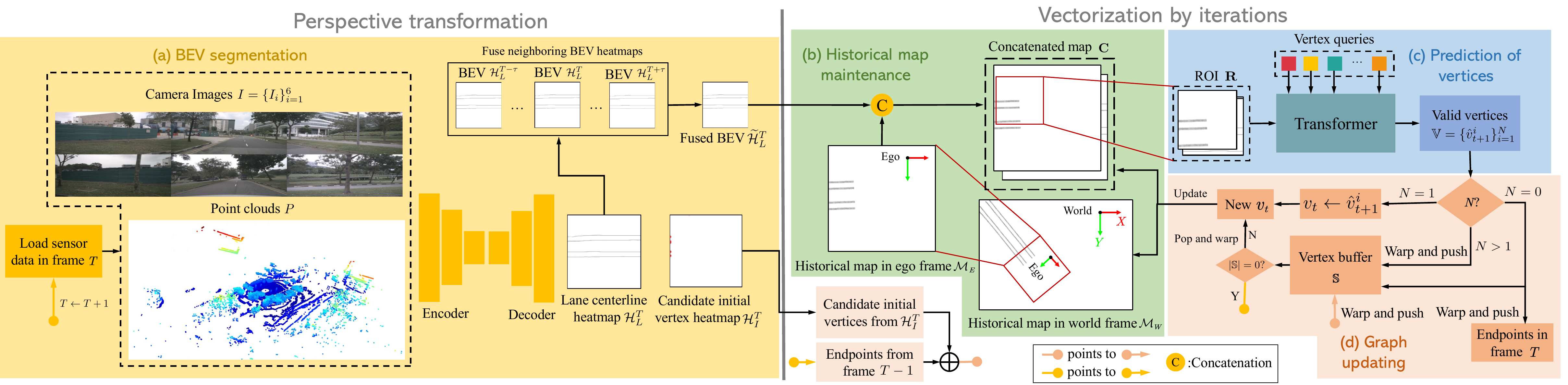}
  \caption{CenterLineDet overview. Our proposed approach consists of two parts: perspective transformation and iterative vectorization. The first part relies on perspective transformation networks (HDMapNet/FusionNet in our work) to predict BEV heatmaps. The second part is the main contribution of this work, which trains a DETR-like transformer network to control an agent to explore the scene by iterations, the trajectory of which is the global vectorized lane centerline graph.}
  \label{diagram}
\end{figure*}

In this work, we aim to detect the road lane centerline graph for HD map automatic generation by using sequential vehicle-mounted sensor data. The input data of our system is a sequence of RGB images captured by six cameras (i.e, $I=\{I_i\}_{i=1}^6$) and a sequence of point clouds obtained by a LiDAR (i.e., $P$). There are multiple frames in the data sequence, and $T$ denotes the current frame. The expected output is the global graph of road lane centerlines in the world coordinate system (i.e.,$G=(V,E)$), where $V$ is a set of lane centerline vertices, and $E$ represents lane centerline edges connecting corresponding adjacent vertices. Fig. \ref{diagram} shows our approach overview.

CenterLineDet has two major steps: In the current frame $T$, (1) predict the BEV heatmap of lane centerlines $\mathcal{H}_L^T$ by perspective transformation, and (2) obtain the lane centerline graph in the world coordinate system. After processing all frames in the input sequence, the expected road lane centerline graph is obtained. For the first step, we propose a network named as FusionNet which enhances the original HDMapNet \cite{li2021hdmapnet} by combining the fully-connected neural view transformer with inverse perspective mapping (IPM), which can extract the lane centerline with better geometric accuracy. Besides $\mathcal{H}_L^T$, we also predict the heatmap of candidate initial vertices $\mathcal{H}_I^T$ by FusionNet. To smooth the inconsistency between frames, we fuse the predicted BEV maps of neighboring frames by warping and averaging. The fused BEV map of lane centerlines at the current time step is denoted as $\tilde{\mathcal{H}}_L^T$.

For the second step, we propose a DETR-like transformer as the decision-making network to control an agent to generate the lane centerline graph vertex by vertex. To start the iteration, we use local peaks in $\mathcal{H}_I^T$ and endpoints in the previous frame $T-1$ as candidate initial vertices of the current frame $T$, which is denoted by $\mathbb{S}=\{s_k\}_{k=1}^K$. 
Vertex $v_t$ is used to denote the current position of the agent. After concatenating the fused BEV segmentation heatmap $\tilde{\mathcal{H}}_L^T$ with the ego historical map $\mathcal{M}_E$, an ROI $\textbf{R}$ is cropped centering on $v_t$ as the local visual information for the agent to make decisions. $\mathcal{M}_E$ is a binary map recording the historical trajectory of the agent in the ego vehicle coordinate system. Taken as input $\textbf{R}$, the transformer predicts $N$ valid vertices in the next step as a set $\mathbb{V}=\{v_{t+1}^i\}_{i=1}^N$. Based on $N$, the agent will take different actions to update the graph iteratively. When the detection of the current lane centerline instance is completed, the agent turns to another candidate initial vertex $s_k$ and repeats the aforementioned algorithm. Once $\mathbb{S}$ is empty, we switch to process the next frame of the sequence. In the end, a graph in the world coordinate system is obtained as the predicted HD map of road lane centerlines.

More details of CenterLineDet are provided in our supplementary document.

\subsection{Perspective Transformation}
To facilitate our centerline graph detection, we first convert the front-view scene from vehicle-mounted sensors to BEV. The BEV centers on the ego vehicle, and its x-axis aligns with the heading direction of the vehicle.

HDMapNet \cite{li2021hdmapnet} applies neural view transformers to transform the feature map $\textbf{F}_{I_i},i\in(1,...,6)$ from each perspective image into the local BEV using fully-connected layers. Then it aligns the local BEV feature map with the global BEV feature map according to the extrinsic parameters of each camera. The mapping $\phi_i$ between the perspective view feature and the BEV feature can be denoted as:
\begin{equation}
    \textbf{F}_{bev}^i = \phi_i(\textbf{F}_{I_i}),
\end{equation}
where $i$ is the index of the camera. In order to improve the generalization ability and the geometric precision of the feature transformation, we propose to enhance the neural view transformer with inverse perspective mapping (IPM). Based on the projective geometry of the camera, IPM computes a mapping between points in the BEV and the perspective view, and obtains a BEV feature map with reliable geometric priors. The FusionNet we proposed treats IPM as a shortcut without learnable parameters and $\phi_i$ as a learnable residual mapping function. The fused camera BEV feature map is the summation of the two mapping results:
\begin{equation}
    \textbf{F}_{bev} = \underset{i}{\text{max}} \left\{ \text{IPM}(\textbf{F}_{I_i}) + \phi_i(\textbf{F}_{I_i}) \right\}.
\end{equation}
The fused BEV features are fed into a sequence of CNN networks to predict a pixel-wise lane centerline segmentation in the BEV.

\subsection{Lane Centerline Graph Detection}
In this section, we show how CenterLineDet detects the lane centerline graph and how the proposed imitation learning algorithm generates expert demonstrations to train the transformer network.

\subsubsection{Inference}
CenterLineDet is trained to mimic expert human annotators to create the HD map of lane centerlines vertex by vertex. It has a DETR-like transformer, a decision-making network controlling an agent to create the HD map of lane centerlines. At each step of the iteration, based on the local visual feature, the agent predicts vertices in the next step and takes corresponding actions to explore the scene. The historical trajectory of the agent is outputted as a prediction of the lane centerline graph. The inference working pipeline of CenterLineDet is shown in Fig. \ref{diagram}. 

To record the historical information which is critical for the decision-making process of CenterLineDet, we maintain a binary historical map $\mathcal{M}_E$. Each frame has a $\mathcal{M}_E$. $\mathcal{M}_E$ is in the ego vehicle coordinate system, which is directly used to guide the decision-making of the agent, while $\mathcal{M}_W$ is in the world coordinate system to assure the consistency of $\mathcal{M}_E$ in neighboring frames. 

At frame $T$ of the input data sequence, after obtaining $\mathcal{H}_L^T$ and $\mathcal{H}_I^T$ from the perspective transformation network, we first find local peak points in $\mathcal{H}_I^T$ and endpoints in the previous frame as a set of candidate initial vertices $\mathbb{S}=\{s_k\}_{k=1}^K$ to initialize the iteration of CenterLineDet. 
Then, starting from a randomly selected $s_k$, CenterLineDet controls an agent to detect one lane centerline instance. Since there exists inconsistency between the BEV segmentation result of different frames, based on ego vehicle poses, we warp and project the neighboring BEV segmentation heatmaps 
$\mathcal{H}_L^{T-\tau} \sim \mathcal{H}_L^{T+\tau}$ to $\mathcal{H}_L^T$. After summation and averaging, the fused BEV heatmap in the current frame $T$ is denoted as $\tilde{\mathcal{H}}_L^T$. Then, we concatenate $\tilde{\mathcal{H}}_L^T$ with $\mathcal{M}_E$. After this, an ROI $\textbf{R}$ centering on the current vertex $v_t$ that the agent locates is cropped, which contains sufficient visual information for the transformer to make the decision. Taken as input $\textbf{R}$, the transformer network outputs the coordinates and valid probability of $\hat{N}$ vertices in the next step $\mathbb{V}=\{v_{t+1}^i\}_{i=1}^{\hat{N}}$. Predicted vertex $v_{t+1}^i$ with high enough valid probability is accepted as a new vertex to update the graph. $\hat{N}$ is the same as the number of input vertex queries. Suppose we have $N$ valid predicted vertices, then the agent should take different actions based on $N$ to handle multiple topology structures of the lane centerline graph. $N=0$ indicates the end of the current lane centerline in the current frame. $v_t$ under such circumstances is treated as an endpoint, which can be a candidate initial vertex in the next frame. The agent should turn to work on another candidate initial vertex in $\mathbb{S}$. $N=1$ means the agent moves along a lane centerline without branches, so the agent should keep moving to the next vertex $v_{t+1}^i$ for graph updating. $N>1$ demonstrates complicated topology structures are met, such as lane intersections, lane splits, and lane merges. The agent should push all $v_{t+1}^i$ into $\mathbb{S}$ as new candidate initial vertices, and pop one $s_k$ from $\mathbb{S}$ to work on.

\begin{figure}[t]
  \centering
  \includegraphics[width=\linewidth]{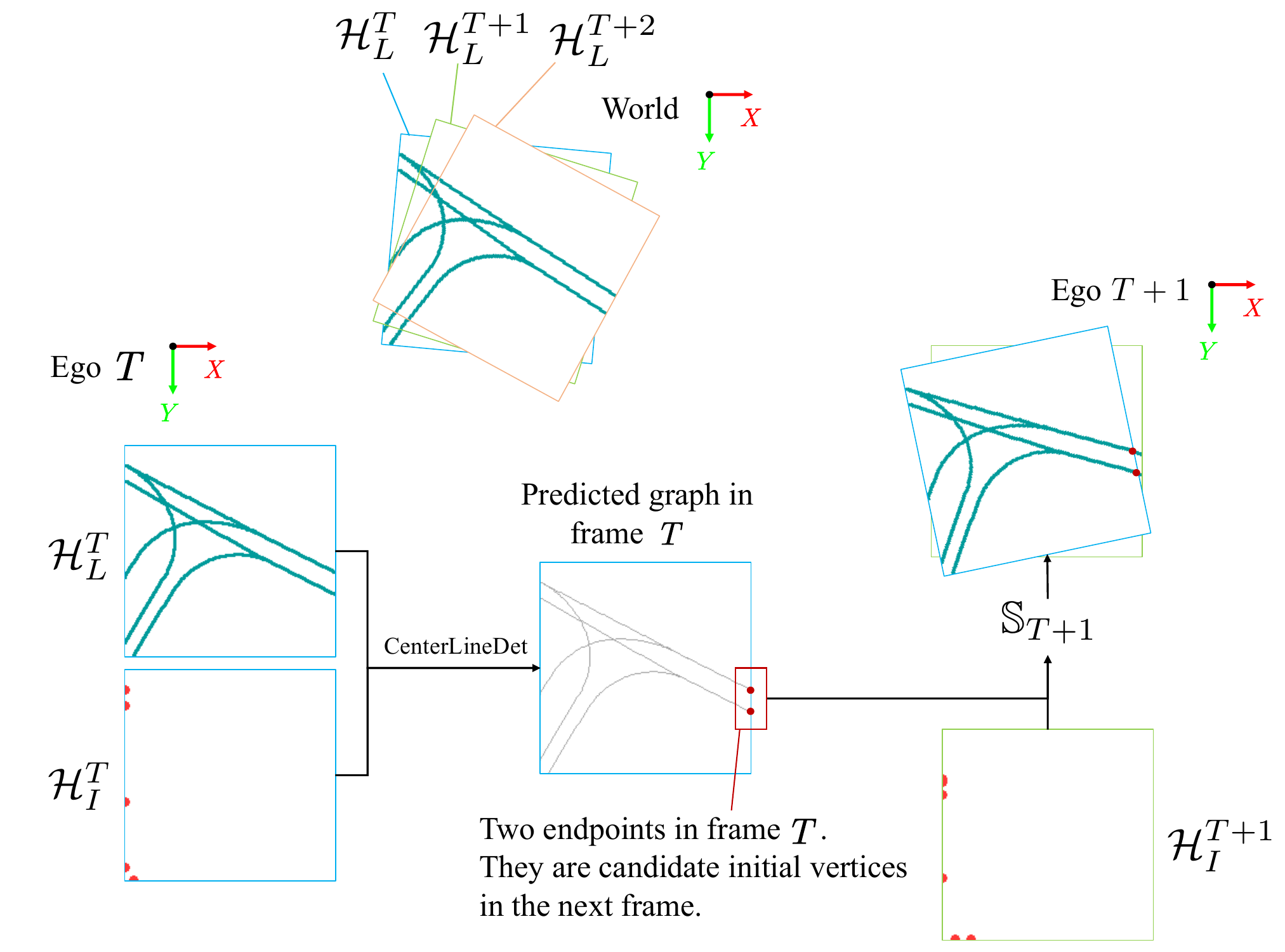}
  \caption{Candidate initial vertices of frame $T+1$. For better visualization, we use the ground truth lane centerline (cyan lines) to represent the fused predicted BEV heatmap $\tilde{\mathcal{H}}_L^T$. The candidate initial vertices of frame $T+1$ come from (1) local peaks of the heatmap $\mathcal{H}_I^{T+1}$ and (2) endpoints of frame $T$. Endpoints are vertices at which the agent decides to stop in the previous frame. This figure is best viewed in color. Please zoom in for details.}
  \label{init}
\end{figure}

When $\mathbb{S}$ is empty, we switch to the next frame $T+1$ of the sequence to continue the detection task. We use endpoints in the current frame (i.e.,$N=0$) and local peaks in $\mathcal{H}_I^{T+1}$ of frame $T+1$ as initialized $\mathbb{S}$ for frame $T+1$. The candidate initial vertices in frame $T$ are visualized in Fig. \ref{init}. For candidate initial vertices that have been explored in the past, the agent ignores them and removes them from $\mathbb{S}$.

After all the frames of the input data sequence are processed, the trajectories of the agent are outputted as the final predicted lane centerline graph.

\subsubsection{Expert Demonstration Sampling}
\begin{figure}[t]
    \includegraphics[width=0.5\textwidth]{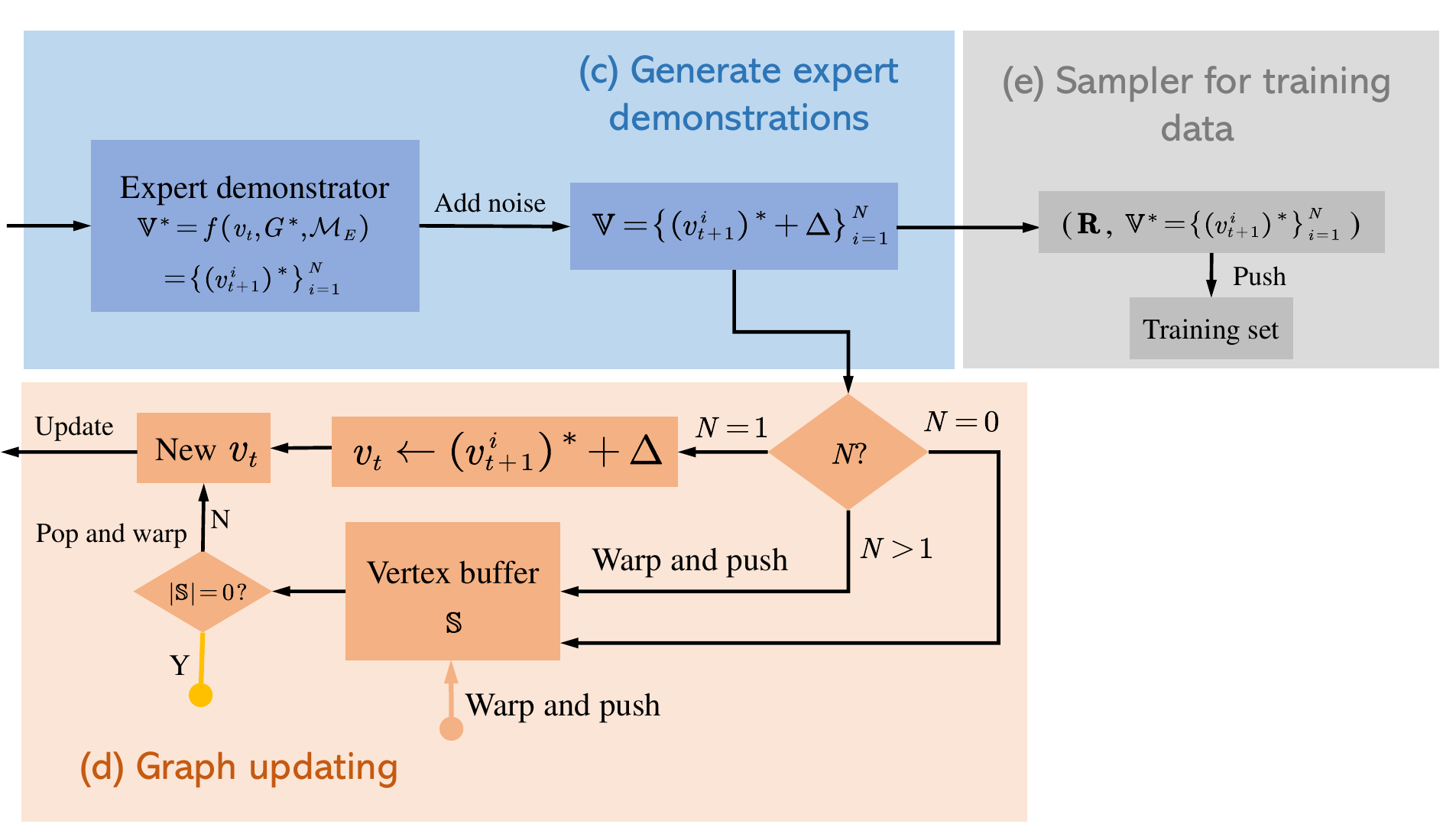}
    \caption{Diagrams of expert demonstration sampling by behavior cloning. For simplicity, in this figure, we only visualize modules (c)-(e), while other modules of this figure are the same as that of Fig. \ref{diagram}. }
    \label{expert}
\end{figure}

In our experiments, training data is generated by a proposed sampling algorithm (i.e., expert in imitation learning). For better training efficiency, in our experiment, behavior-cloning sampling algorithm \cite{osa2018algorithmic} is adopted. Based on breath-first-search (BFS), the sampling algorithm traversals the ground truth lane centerline graph $G^*$ vertex by vertex. At each position $v_t$, it generates one training sample. To enhance the robustness of CenterLineDet, we add evenly distributed noises to the expert trajectory during data sampling. The simplified diagram of the sampling algorithm is visualized in Fig. \ref{expert}.

During the sampling of the behavior-cloning algorithm, the ground truth label of the current step is obtained by the following equation:
\begin{equation}
    \mathbb{V}^*=f(v_t,G^*,M_E),
\end{equation}
where function $f(\cdot)$ can calculate the label vertices in the next step based on the ground truth graph $G^*$ and historical information. In our experiments, $f(\cdot)$ is modified from the labeling algorithm proposed in \cite{xu2022rngdet}, which can handle the labeling task in graphs with complicated topology. To make CenterLineDet more robust, we add random noises to $\mathbb{V}^*$ when updating the graph. 

\section{Experimental Results and Discussions}
\label{sec:result}
To evaluate and verify the superiority of our proposed CenterLineDet, we conduct comparative experiments and ablation studies on the public NuScenes dataset \cite{caesar2020nuscenes}. NuScenes is a large dataset containing data collected from various different autonomous driving scenarios. This dataset provides hundreds of data sequences collected by vehicle-mounted sensors. Each sequence has around 40 frames with a 2Hz frame rate. We select 698 sequences for training, and 148 sequences for inference. Since CenterLineDet is of two stages and heavily relies on the BEV segmentation of perspective transformation, scene sequences that either have no centerlines or perspective transformation has no reasonable outputs are not included in the inference set.

\subsection{Evaluation Metrics}
To evaluate the performance of approaches from both pixel-level and topology-level, we modify the metrics used in \cite{he2022lane} and \cite{he2020sat2graph} for our experiments. 
There are three pixel-level metric scores, pixel-precision (P-Pre), pixel-recall (P-Rec) and pixel-f1 (P-F1), to evaluate the prediction correctness at the pixel scale. Suppose we have the predicted graph $\hat{G}$ and ground truth graph $G^*$. For a vertex $p$ in $\hat{G}$, if there exists one vertex $q$ in $G^*$ whose distance to $p$ is smaller than a threshold $\delta$, then $p$ is regarded as a correct prediction. Similarly, for a vertex $q$ in $G^*$, if there exists one vertex $p$ in $\hat{G}$ whose distance to $q$ is smaller than $\delta$, then $q$ is correctly retrieved. The pixel-level metrics can be calculated based on the following equations:
\begin{equation}
\begin{aligned}
    \text{P-Pre}&=\frac{|\{p|\lVert p-q \rVert<\delta,p\in \hat{G},\exists q \in G^*\}|}{|\hat{G}|}\\
    \text{P-Rec}&=\frac{|\{q|\lVert p-q \rVert<\delta,q\in G^*,\exists p \in \hat{G}\}|}{|G^*|},
\end{aligned}
\end{equation}
where $|\cdot|$ is the cardinality of a set. P-F1 is a combination of P-Pre and P-Rec, which is equal to $\frac{2\text{P-Pre}\cdot \text{P-Rec}}{\text{P-Pre}+\text{P-Rec}}$.

There are also three metrics to evaluate the topology correctness of the predicted graph, that is, topology-precision (T-Pre), topology-recall (T-Rec) and topology-f1 (T-F1). For each vertex $q$ in $G^*$, we find that all vertices in $G^*$ that $q$ can reach within distance $\epsilon$ as a sub-graph $G^*_{q}$. Then, we find the vertex in $\hat{G}$ that is closest to $q$ as $\tilde{p}$. The sub-graph $\hat{G}_{\tilde{p}}$ denotes all vertices in $\hat{G}$ that $\tilde{p}$ can reach, whose distance to $\tilde{p}$ is smaller than $\epsilon'$. 
After calculating pixel-level scores between the obtained sub-graphs $G^*_{q}$ and $\hat{G}_{\tilde{p}}$, we have the topology-scores:
\begin{equation}
    \text{T-X}=\frac{\sum_{q\in G^*}\text{P-X}(G^*_{q},\hat{G}_{\tilde{p}})}{|G^*|},
\end{equation}
where $\tilde{p}$ is the closest point in $\hat{G}$ to $q$, and the letter X can be Pre, Rec or F1.

\subsection{Comparative Results}
In this section, we evaluate CenterLineDet under different settings and compare our CenterLineDet with baseline models. 
To ensure fair and comprehensive comparisons, we evaluate all approaches in both single-frame and multi-frame detection tasks. Single-frame and multi-frame evaluation results are shown in Tab. \ref{tab_comp_multi}. Sample qualitative results are visualized in Fig. \ref{vis}. Since CenterLineDet does not aim for the single-frame task, we only report the pixel-level results of single frames.

\begin{table}[t] 
\setlength{\abovecaptionskip}{0pt} 
\setlength{\belowcaptionskip}{0pt} 
\renewcommand\arraystretch{1.0} 
\renewcommand\tabcolsep{0.8pt} 
\centering 
\begin{threeparttable}
\caption{The quantitative results for comparison experiments. }
\begin{tabular}{@{}c c c c c c c c c c c c c c c c c c c c c c c@{}}
\toprule
 \multirow{5}{*}{Approaches}& \multicolumn{3}{c}{Single-frame} & \multicolumn{6}{c}{Multi-frame}\\ 
\cmidrule(l){2-4} \cmidrule(l){5-10} 
&\multicolumn{3}{c}{Pixel-level $\uparrow$} & \multicolumn{3}{c}{Pixel-level $\uparrow$} & \multicolumn{3}{c}{Topology-level $\uparrow$}\\
\cmidrule(l){2-4} \cmidrule(l){5-7} \cmidrule(l){8-10} 
 &   P-Pre &  P-Rec &  P-F1 &  P-Pre &  P-Rec &  P-F1 &   T-Pre & T-Rec & T-F1& \\
\midrule
HDMapNet \cite{li2021hdmapnet}  & 0.805 & 0.649 & 0.709 &0.714&0.665&0.685&0.517 & 0.354 & 0.400 \\
TopoRoad \cite{can2021topology} & 0.408 & 0.566 & 0.461  & 0.410&0.526&0.477&0.360&0.349&0.352 \\
FusionNet  & \textbf{0.813} & 0.658 & 0.719 &0.726&0.674&0.695&0.518 & 0.356 & 0.403\\
\midrule
CenterLineDet\\
\qquad+HDMapNet & 0.785 & 0.675 & 0.711 & 0.700 & \textbf{0.713} & 0.699 & 0.768 & 0.403 & 0.511 \\
\qquad+FusionNet &0.811&\textbf{0.675}&\textbf{0.725}& \textbf{0.732}&0.708&\textbf{0.714}&\textbf{0.782}&\textbf{0.409}&\textbf{0.518} \\
\bottomrule 
\label{tab_comp_multi}
\end{tabular} 
\end{threeparttable}
\end{table}

\subsubsection{Baselines} 
To the best of our knowledge, very few past works have exactly the same research scope as ours, that is, detecting the graph of road lane centerlines in sequential data collected by vehicle-mounted sensors. They either only focus on single-frame detection tasks or resort to other format input data (e.g., aerial image and OSM). Therefore, in the comparison experiments, we create our own baseline models based on past works:
\begin{itemize}
    \item HDMapNet \cite{li2021hdmapnet} (ICRA2022):
    HDMapNet is one of the state-of-the-art approaches for road element perception and vectorization with perspective transformation, but it is mainly designed for the single-frame detection task. To adapt it to our multi-frame detection task, we apply the frame merging method proposed in \cite{elhousni2020automatic} to construct the world-level graph of lane centerlines.
    \item TopoRoad \cite{can2021topology} (CVPR2022):
    TopoRoad is a DETR-like model, which can predict lane centerlines as B-splines in a single frame. We apply the same frame merging method as the HDMapNet baseline to merge multiple frames in the data sequence. 
    \item FusionNet: FusionNet is proposed in this paper, which enhances HDMapNet by combining IPM with fully-connected neural view transformers to better learn the geometric transformation. Similar to the HDMapNet baseline, we first obtain single-frame heatmaps and then merge them into the final multi-frame detection result.
\end{itemize}

\subsubsection{CenterLineDet} 
We evaluate and show the results of CenterLineDet with different perspective transformation networks (i.e., HDMapNet and FusionNet). CenterLineDet is trained by behavior-cloning.

 \begin{figure*}[!t]
    \begin{subfigure}[t]{0.195\textwidth}
        \begin{subfigure}[t]{\textwidth}
            \includegraphics[width=\textwidth]{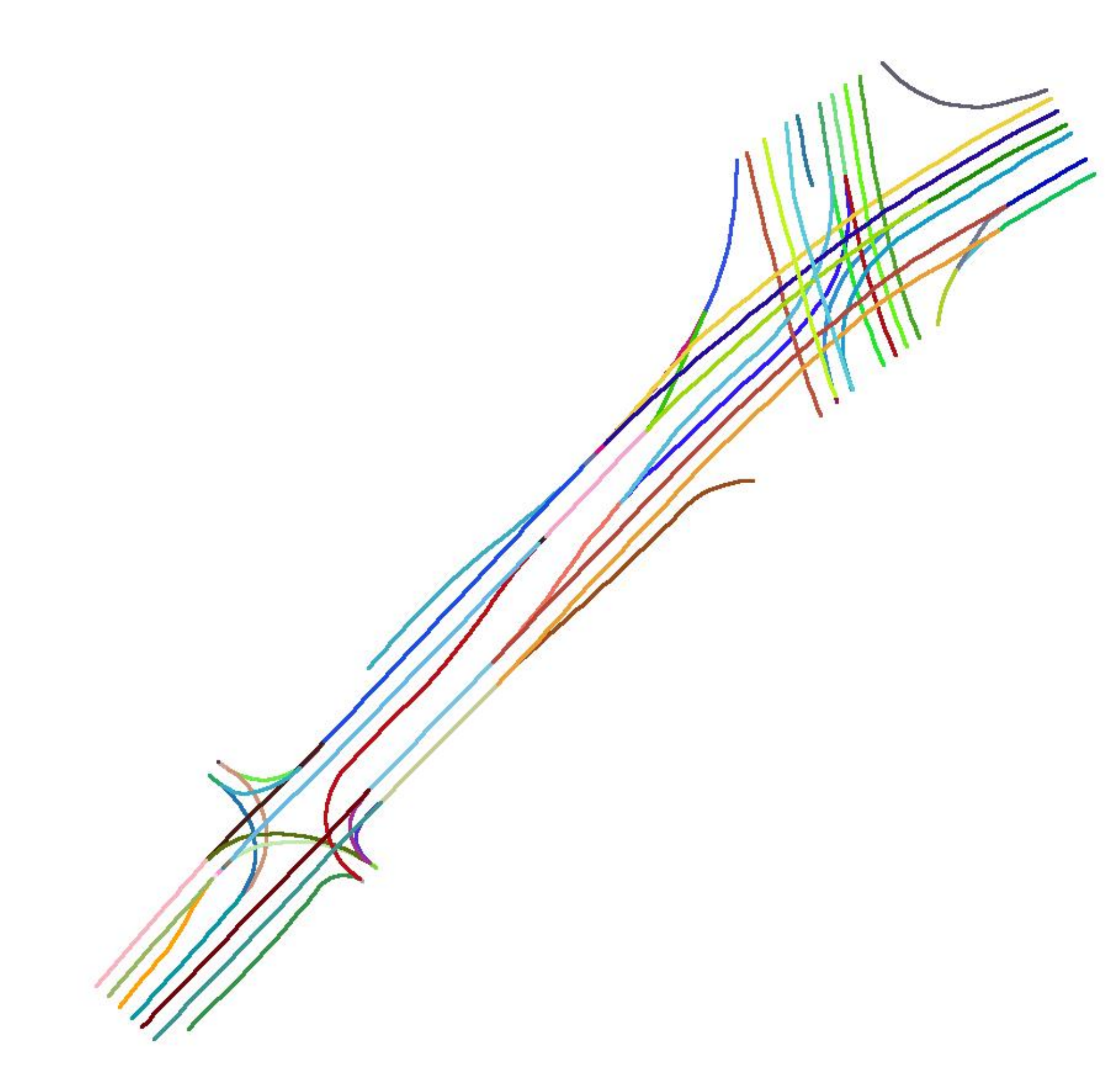}
        \end{subfigure}\vspace{.6ex}
        \begin{subfigure}[t]{\textwidth}
            \includegraphics[width=\textwidth]{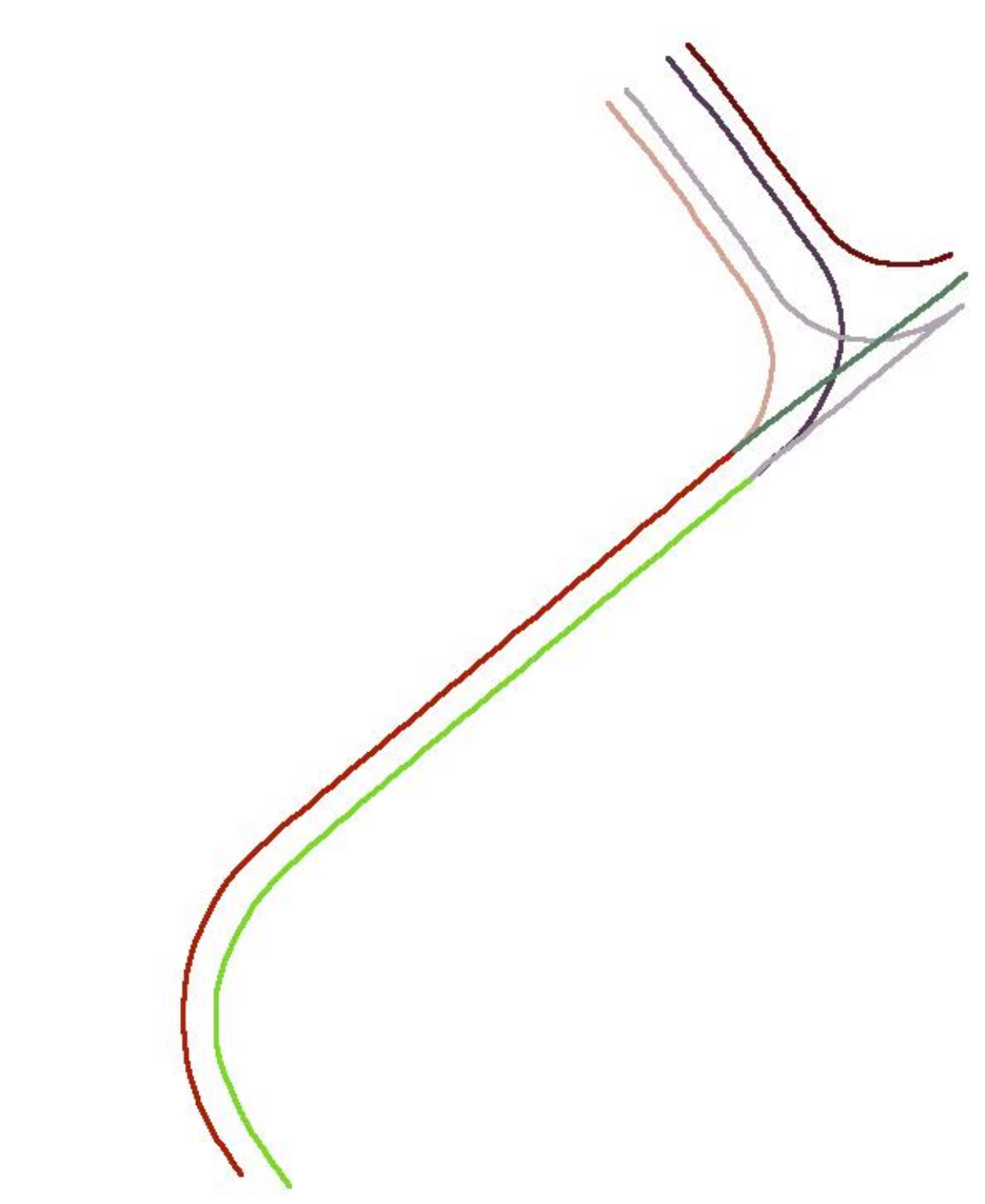}
        \end{subfigure}\vspace{.6ex}
        \begin{subfigure}[t]{\textwidth}
            \includegraphics[width=\textwidth]{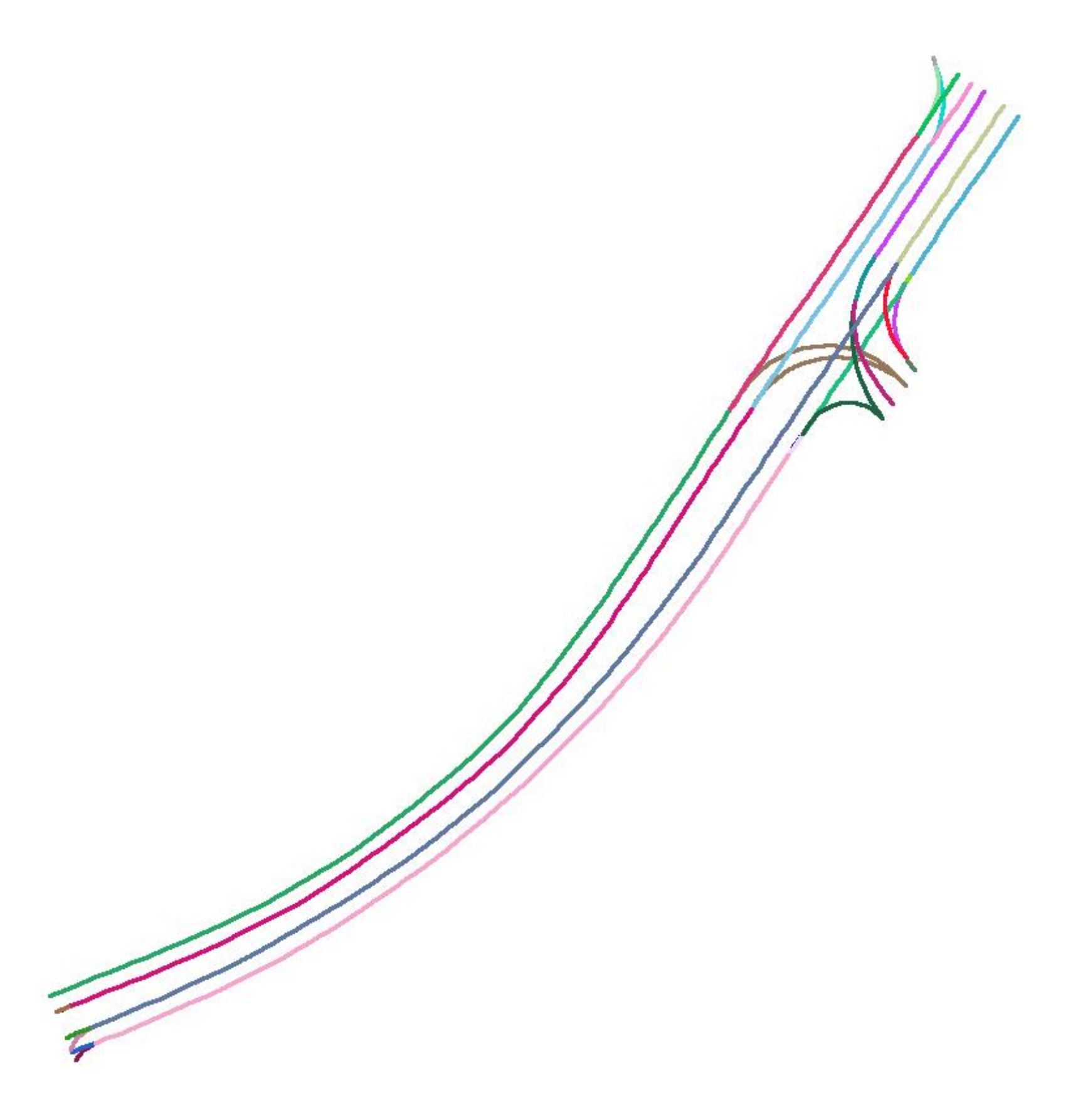}
        \end{subfigure}\vspace{.6ex}
        \caption{GT}
    \end{subfigure}
    \begin{subfigure}[t]{0.195\textwidth}
        \begin{subfigure}[t]{\textwidth}
            \includegraphics[width=\textwidth]{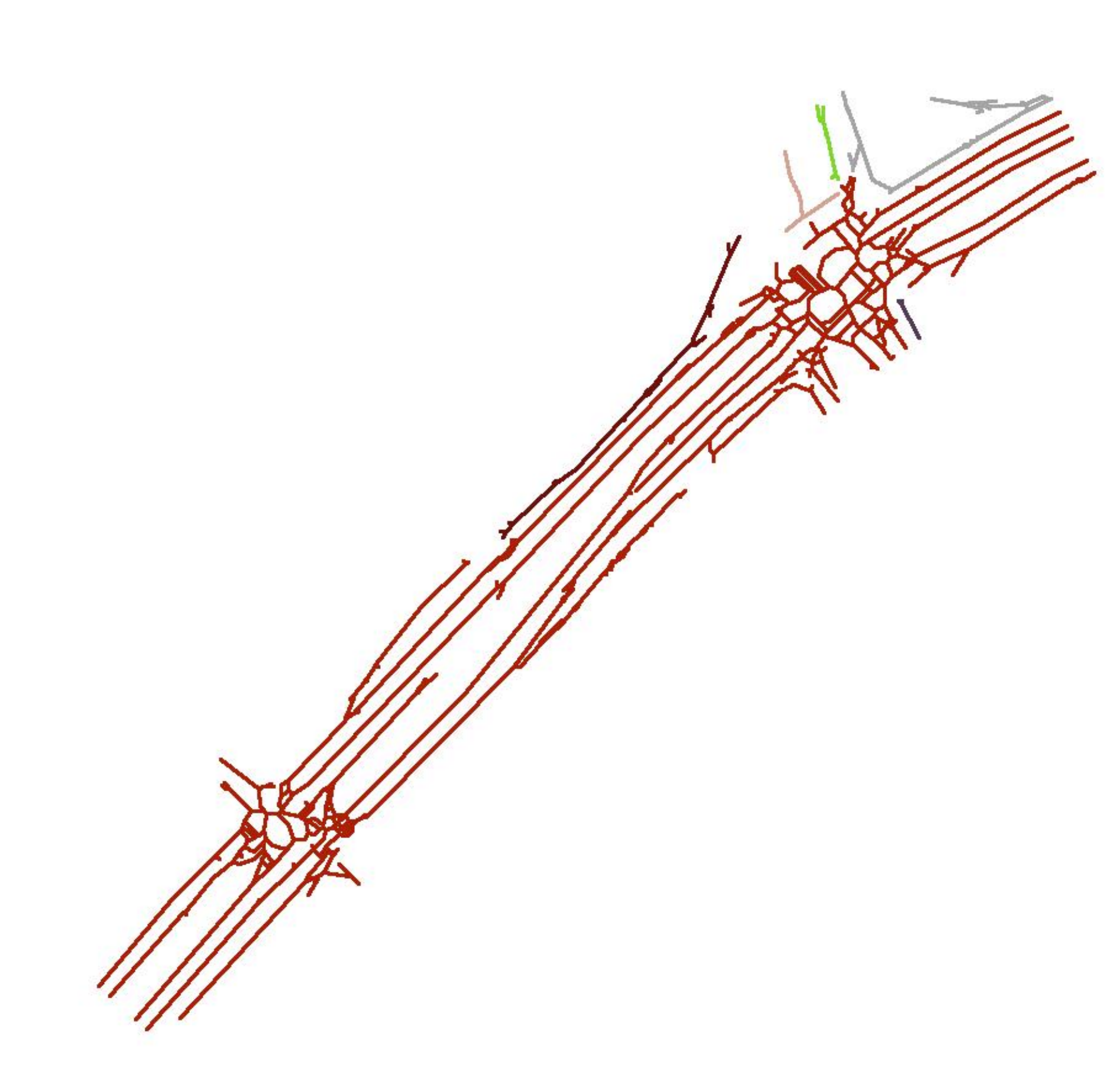}
        \end{subfigure}\vspace{.6ex}
        \begin{subfigure}[t]{\textwidth}
            \includegraphics[width=\textwidth]{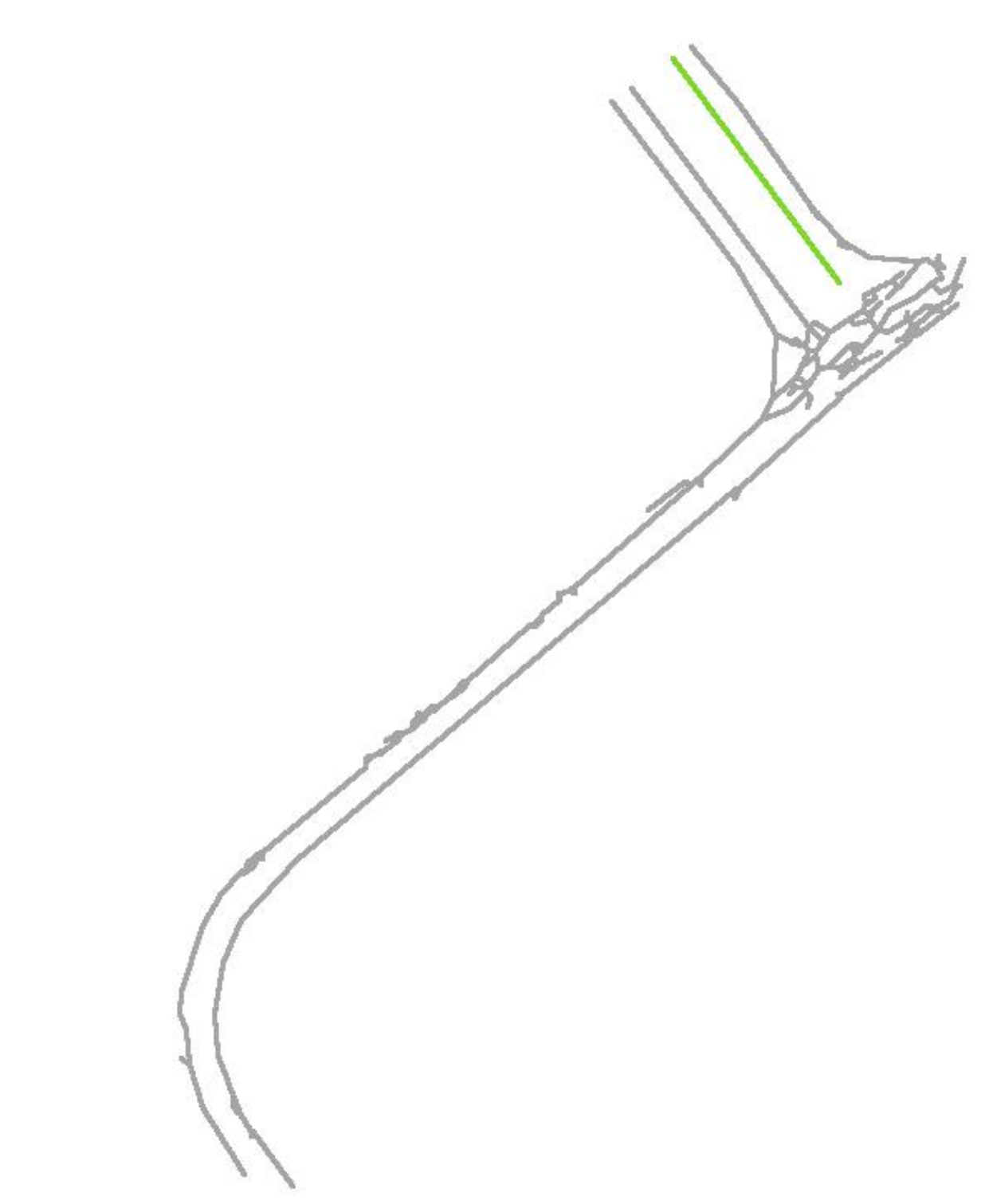}
        \end{subfigure}\vspace{.6ex}
        \begin{subfigure}[t]{\textwidth}
            \includegraphics[width=\textwidth]{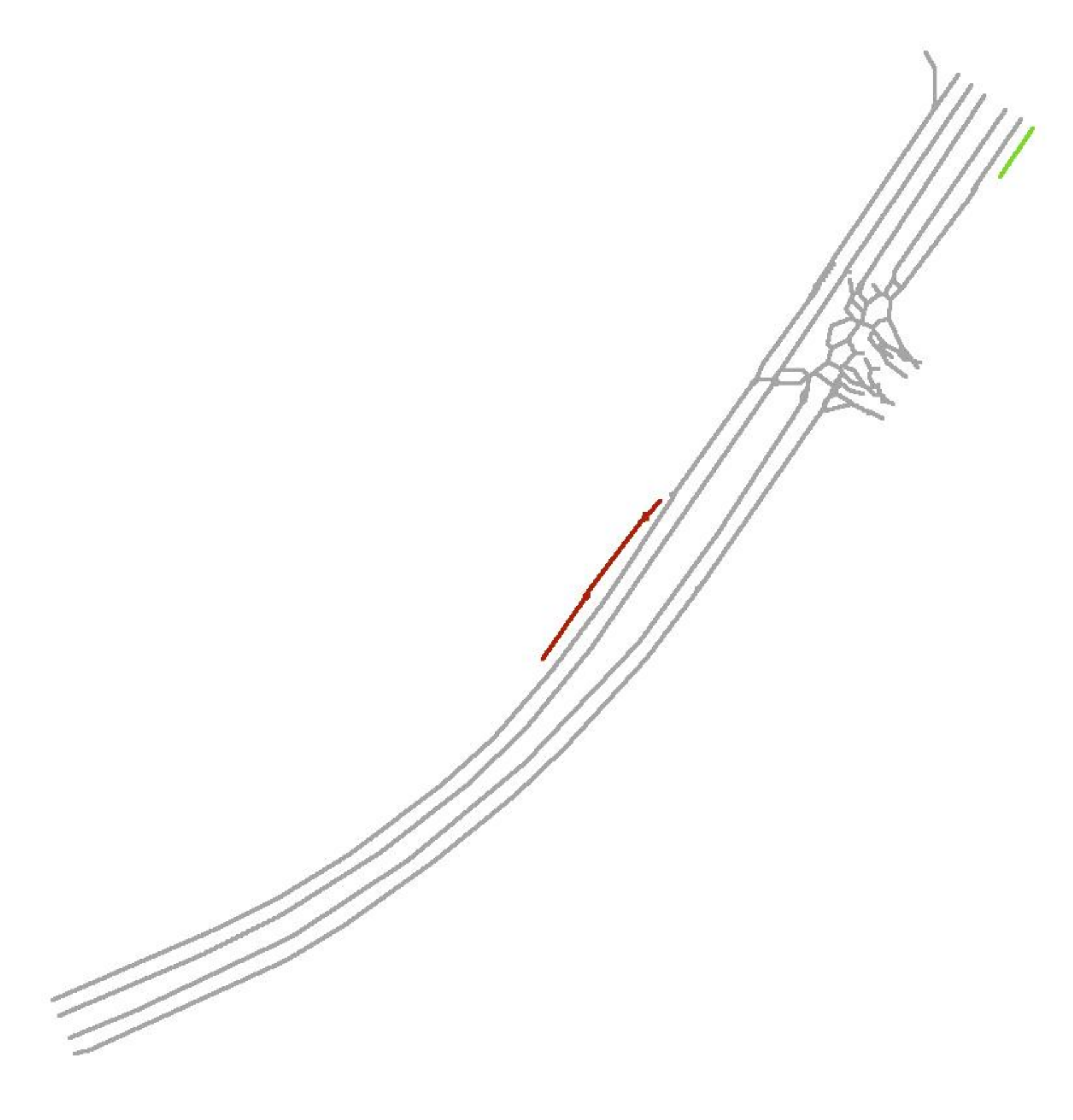}
        \end{subfigure}\vspace{.6ex}
        \caption{HDMapNet \cite{li2021hdmapnet}}
    \end{subfigure}
    \begin{subfigure}[t]{0.195\textwidth}
        \begin{subfigure}[t]{\textwidth}
            \includegraphics[width=\textwidth]{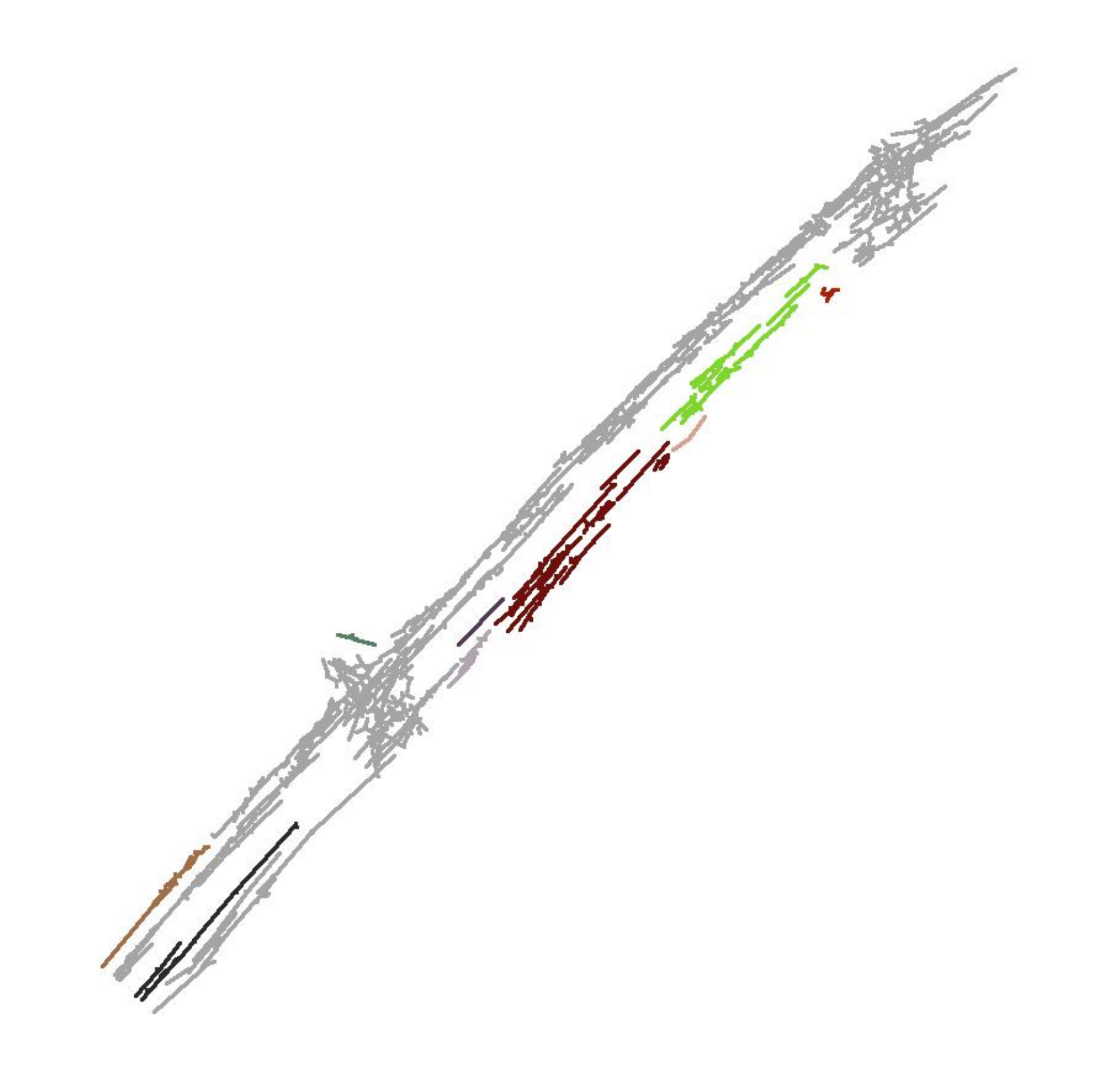}
        \end{subfigure}\vspace{.6ex}
        \begin{subfigure}[t]{\textwidth}
            \includegraphics[width=\textwidth]{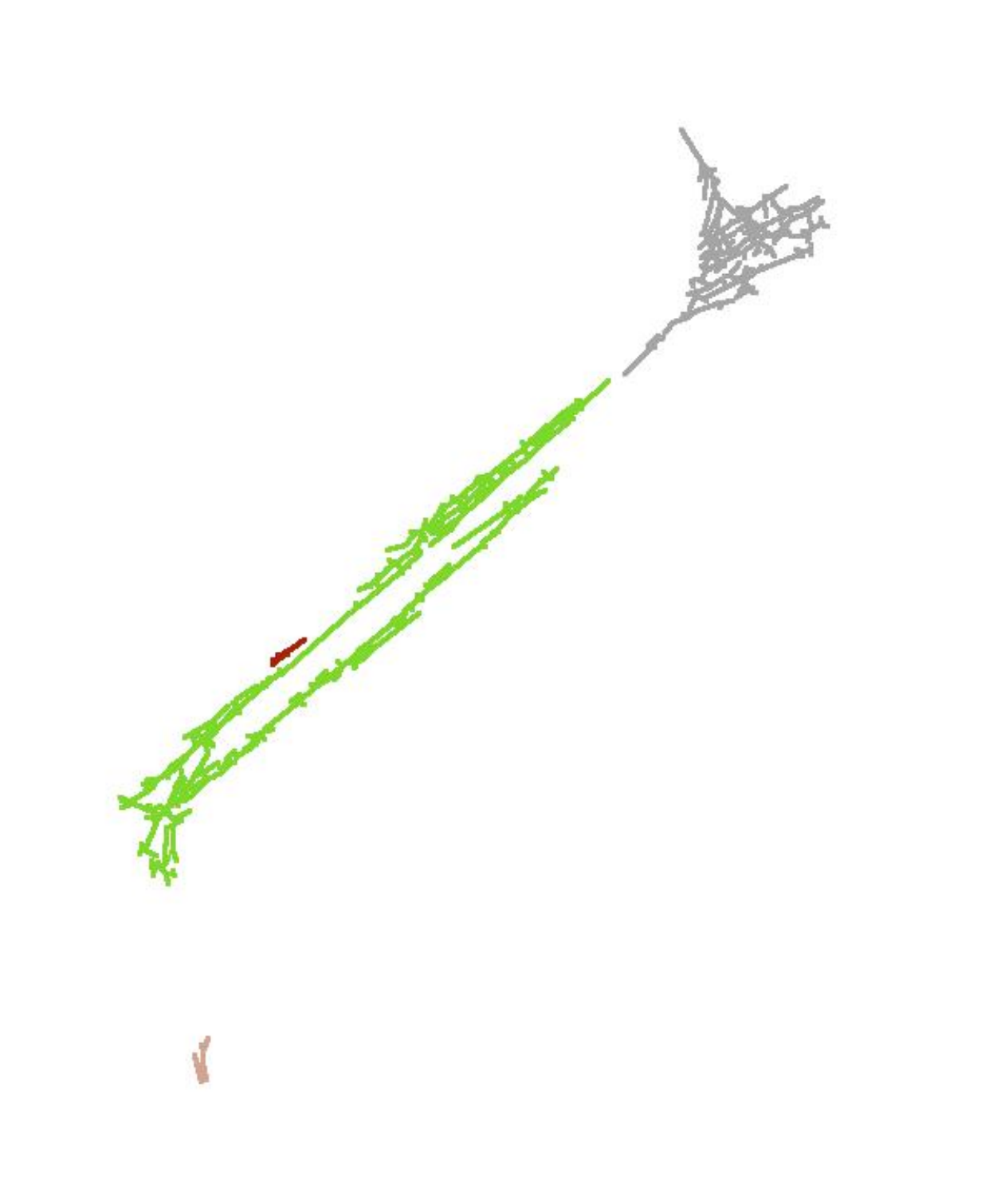}
        \end{subfigure}\vspace{.6ex}
        \begin{subfigure}[t]{\textwidth}
            \includegraphics[width=\textwidth]{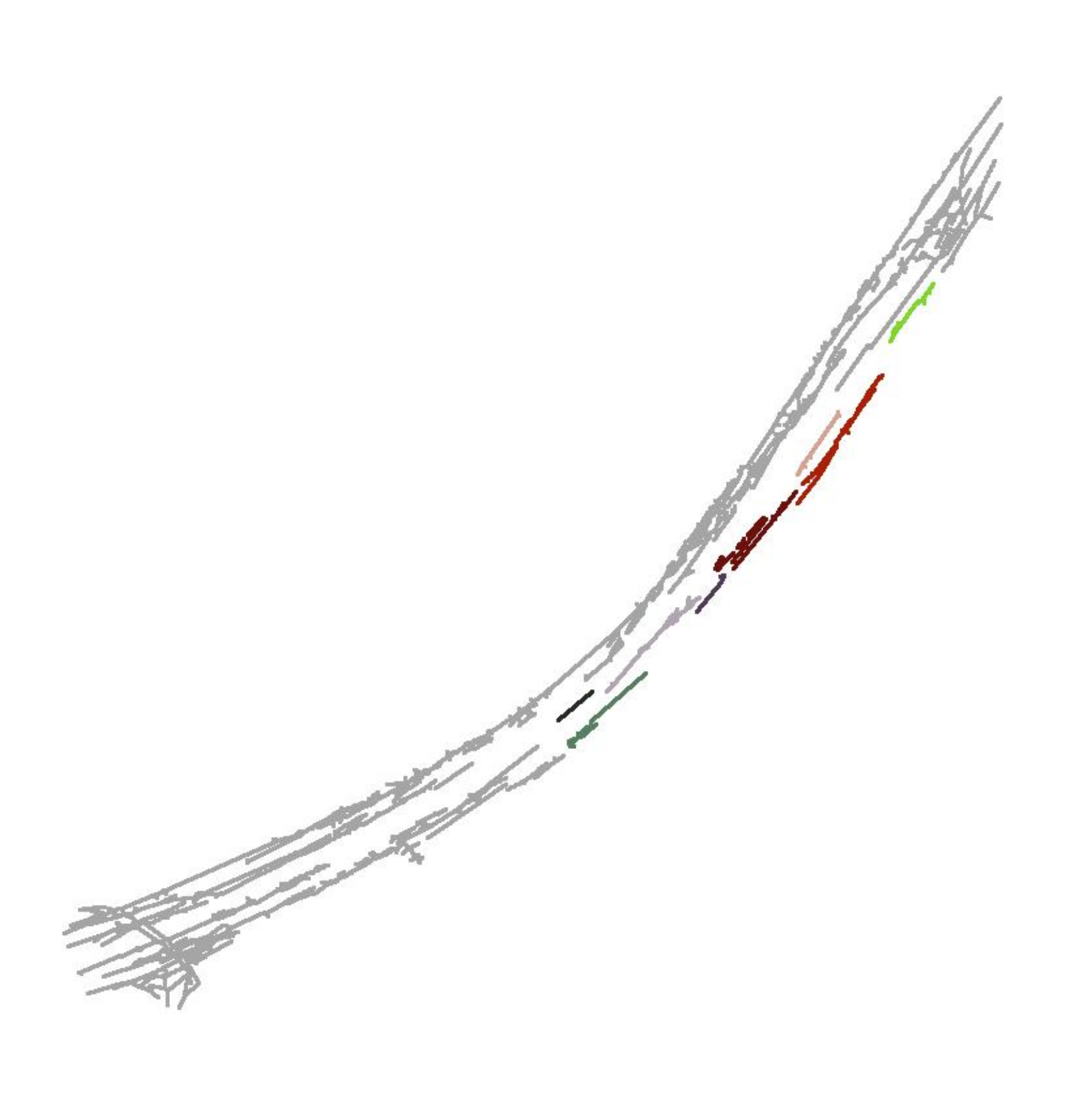}
        \end{subfigure}\vspace{.6ex}
        \caption{TopoRoad \cite{can2021topology}}
    \end{subfigure}
    \begin{subfigure}[t]{0.195\textwidth}
        \begin{subfigure}[t]{\textwidth}
            \includegraphics[width=\textwidth]{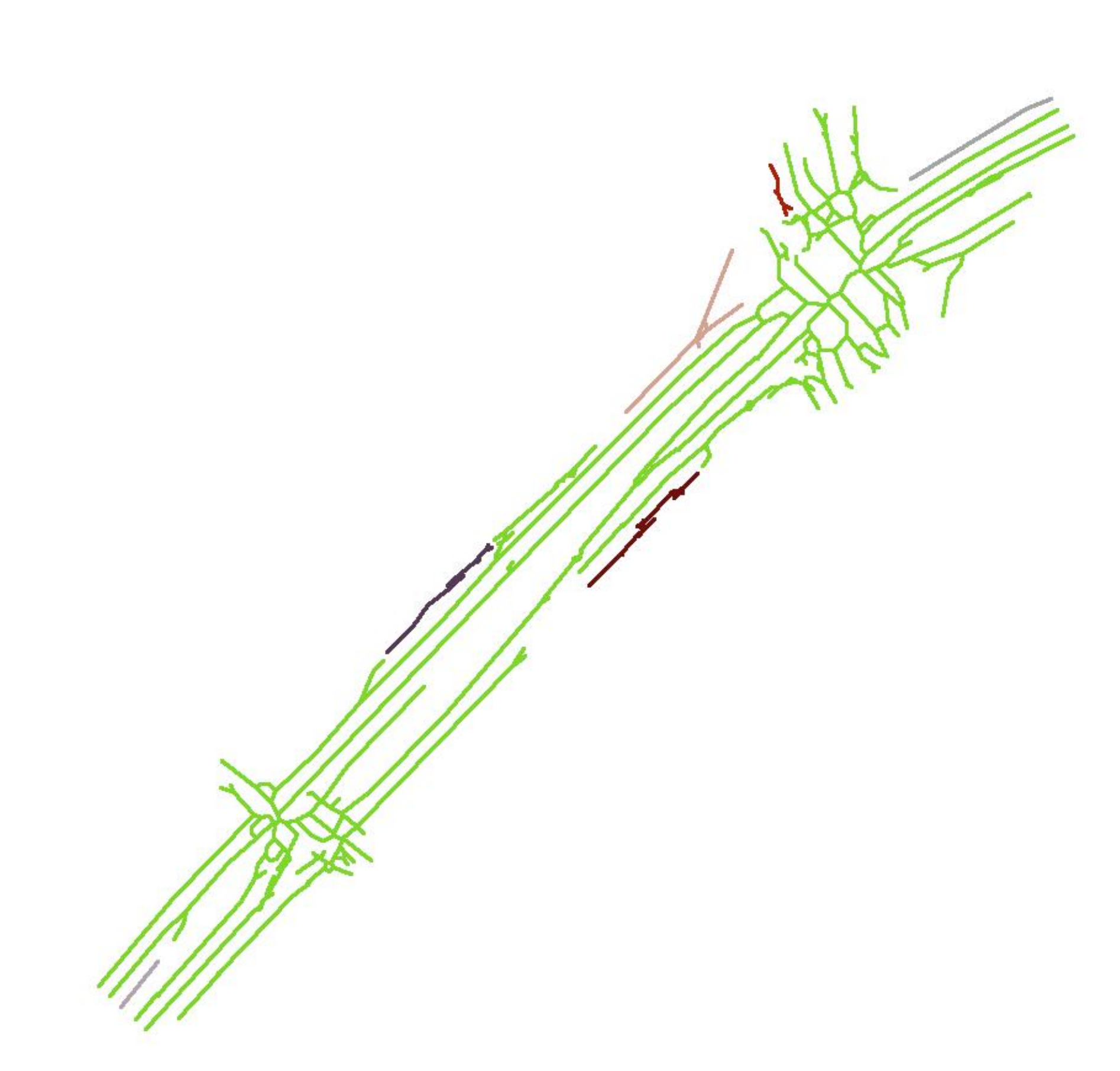}
        \end{subfigure}\vspace{.6ex}
        \begin{subfigure}[t]{\textwidth}
            \includegraphics[width=\textwidth]{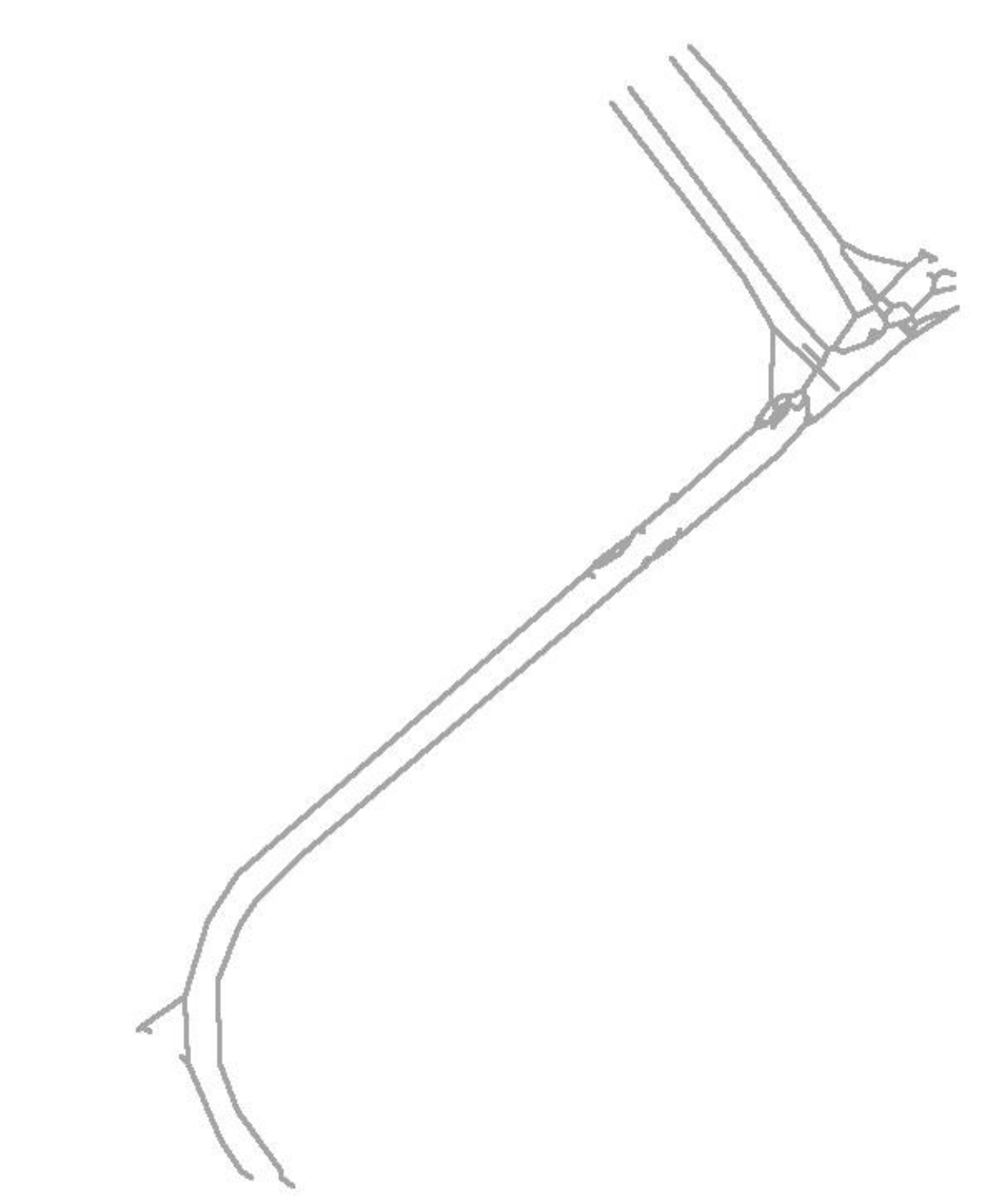}
        \end{subfigure}\vspace{.6ex}
        \begin{subfigure}[t]{\textwidth}
            \includegraphics[width=\textwidth]{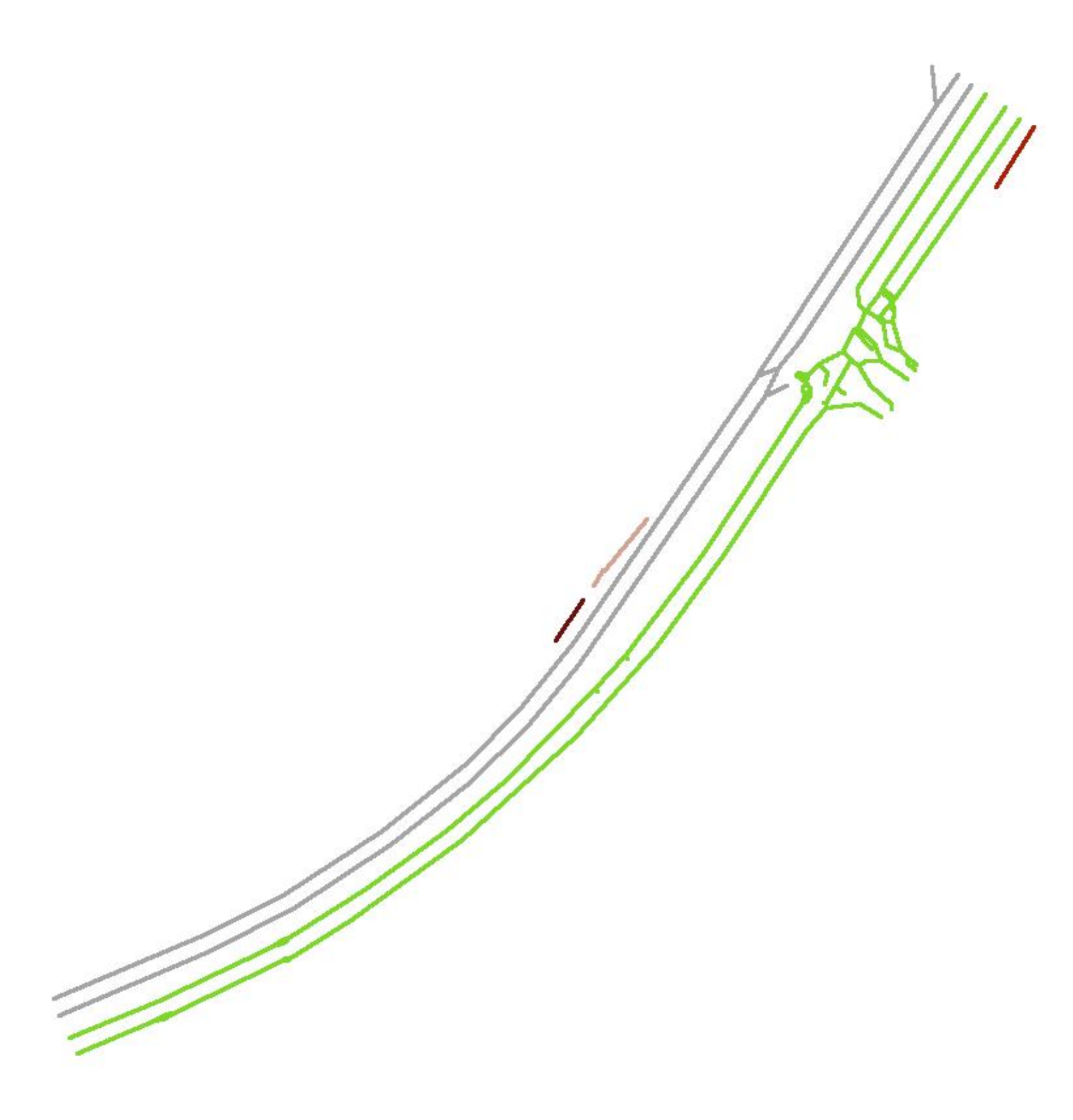}
        \end{subfigure}\vspace{.6ex}
        \caption{ FusionNet}
    \end{subfigure}
    \begin{subfigure}[t]{0.195\textwidth}
        \begin{subfigure}[t]{\textwidth}
            \includegraphics[width=\textwidth]{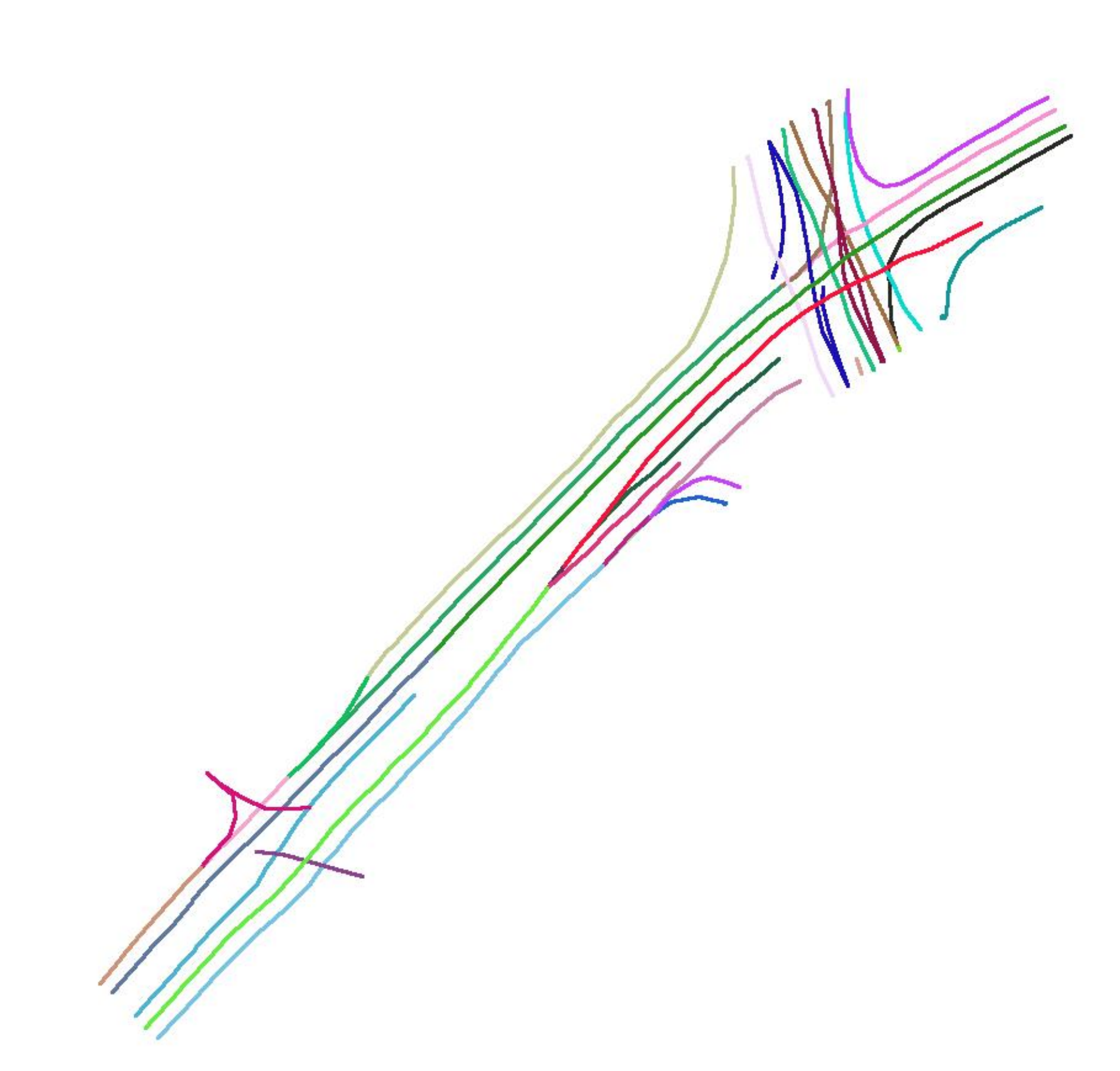}
        \end{subfigure}\vspace{.6ex}
        \begin{subfigure}[t]{\textwidth}
            \includegraphics[width=\textwidth]{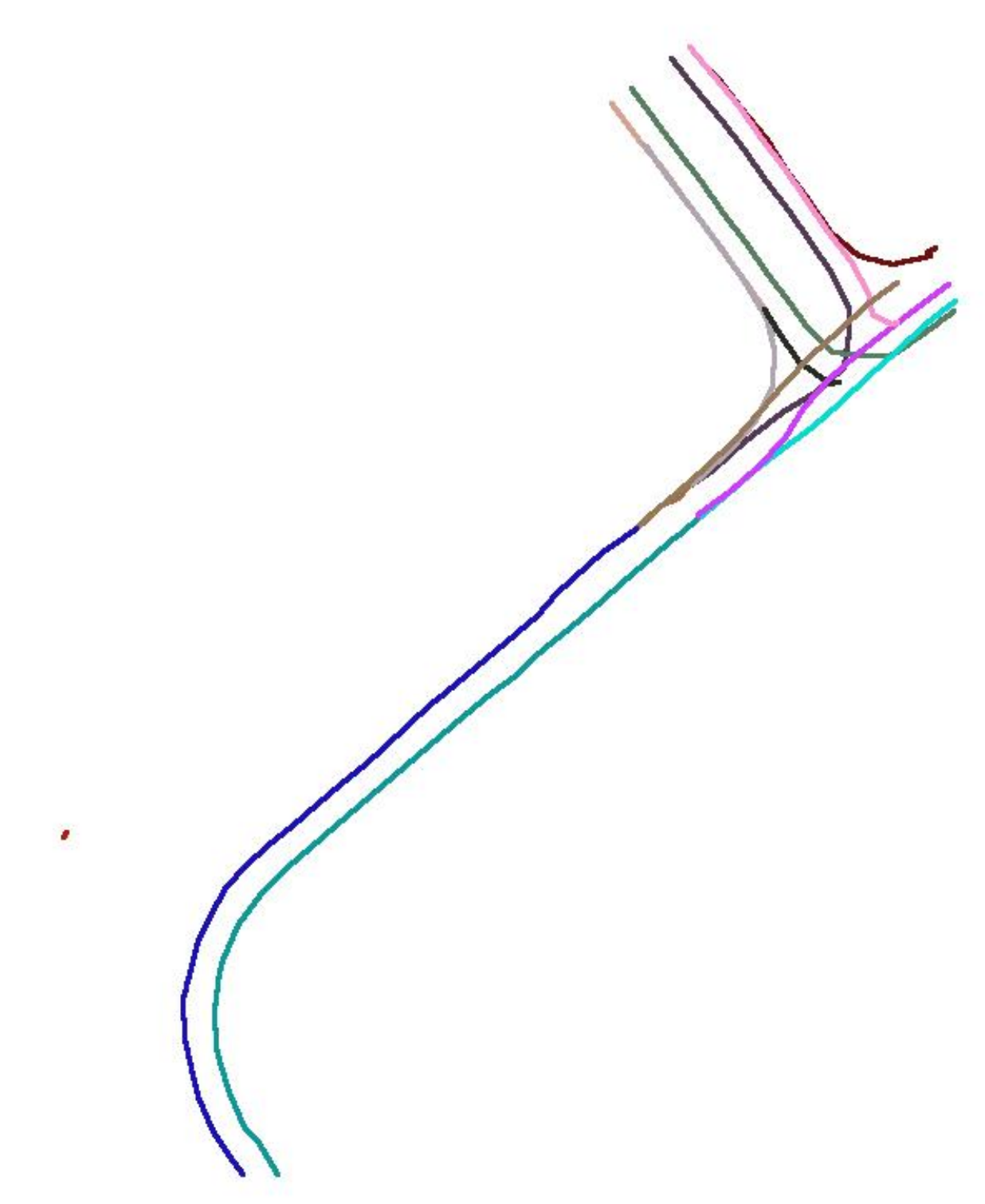}
        \end{subfigure}\vspace{.6ex}
        \begin{subfigure}[t]{\textwidth}
            \includegraphics[width=\textwidth]{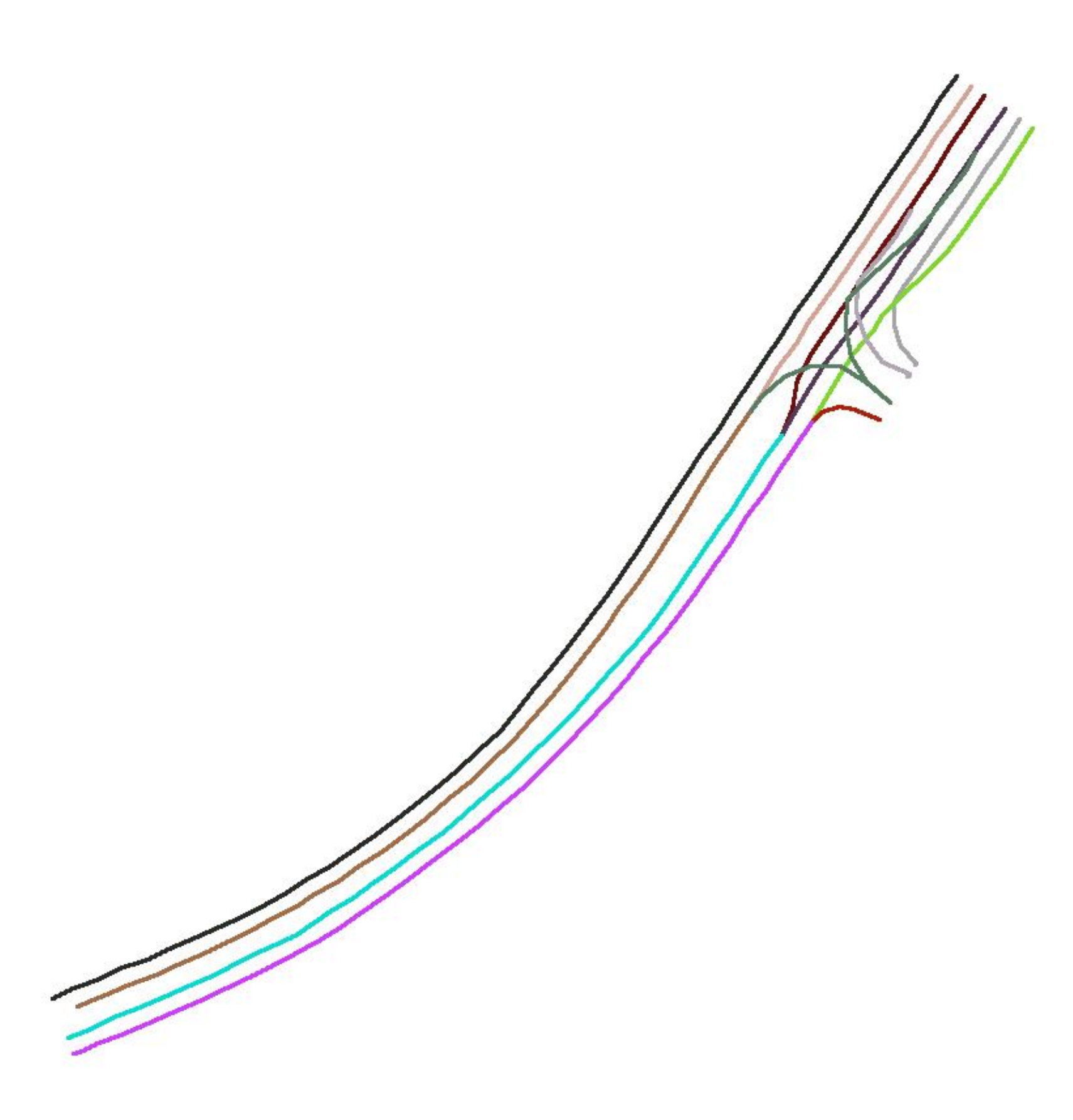}
        \end{subfigure}\vspace{.6ex}
        \caption{ CenterLineDet}
    \end{subfigure}
    \caption{Qualitative visualizations. Different colors represent different road centerline instances. CenterLineDet is the only approach that can detect and distinguish multiple instances. All baselines mess up different instances, which leads to incorrect topology, especially in the intersection area. For better visualization, graphs are widened but they are actually of one-pixel width. Please refer to the supplementary document for additional visualizations. Please zoom in for details.}
    \label{vis}
\end{figure*}

\subsubsection{Discussions} 
TopoRoad \cite{can2021topology} outputs a noisy graph in each frame, and it is almost impossible to merge graphs of each frame into a consistent global one. From the results in Tab. \ref{tab_comp_multi}, we observe that FusionNet gains enhancement than past works, which proves the effectiveness of the fusion of IPM and fully-connected neural view transformers. For each perspective transformation model, the corresponding CenterLineDet presents superior performance, especially in multi-frame evaluations. This is because the agent controlled by CenterLineDet can make more appropriate decisions for graph detection. From visualizations in Fig. \ref{vis}, it is observed that CenterLineDet is the only approach that can distinguish centerline instances, while baselines mess up instances that causes incorrect topology. 

\subsection{Ablation Studies}
 
\begin{table}[t] 
\setlength{\abovecaptionskip}{0pt} 
\setlength{\belowcaptionskip}{0pt} 
\renewcommand\arraystretch{1.0} 
\renewcommand\tabcolsep{3.4pt} 
\centering 
\begin{threeparttable}
\caption{The quantitative results for multi-frame ablation studies.}
\begin{tabular}{@{}c c c c c c c c c c c c c c c c c c c c c c@{}}
\toprule
 \multirow{3}{*}{Approaches}& \multicolumn{3}{c}{Pixel-level $\uparrow$} & \multicolumn{3}{c}{Topology-level $\uparrow$}\\ 
\cmidrule(l){2-4} \cmidrule(l){5-7} 
 &   P-Pre &  P-Rec &  P-F1 &   T-Pre & T-Rec & T-F1 \\
 \midrule
CenterLineDet-No LiDAR  & 0.719 & 0.683 & 0.698 & 0.721 & 0.365 & 0.466\\
CenterLineDet-No Camera & 0.631 & 0.576 & 0.605 & 0.620 & 0.338 & 0.394 \\
\midrule
CenterLineDet   &\textbf{0.732}&\textbf{0.708}&\textbf{0.714} &\textbf{0.782}&\textbf{0.409}&\textbf{0.518} \\
\bottomrule 
\label{tab_ablation}
\end{tabular} 
\end{threeparttable}
\end{table}

In this section, we verify the importance of the input data by respectively removing one of two sensors (i.e., camera and LiDAR). The quantitative results are shown in Tab. \ref{tab_ablation}. We can see that removing either LiDAR or cameras will degrade the results, and our CenterLineDet without cameras has much inferior performance. This indicates the importance of data fusion, and camera images are the dominant source of information for our CenterLineDet. Based on the aforementioned observations, the effectiveness of our network design is demonstrated. 

\subsection{Time cost}
We conduct experiments on a server with an i7-8700K CPU and four RTX-3090 GPUs. All the four GPUs are utilized for training, while only a single GPU is used during inference. We report the time cost as follows:

\begin{itemize}
\item It takes one day to train HDMapNet or FusionNet.
\item It takes 13 minutes to infer 5981 frames (148 scenes) for HDMapNet or FusionNet.
\item It takes around 5 hours for behavior-cloning sampling, and it takes one extra day to train CenterLineDet with behavior-cloning sampled data.
\item It takes overall 41 minutes for CenterLineDet to infer 5981 frames (0.41s/frame=2.43Hz, which is sufficient for Nuscenes with 2Hz key frame rate). Besides, it should be noted that CenterLineDet does not need to work in an online manner (i.e., HD map generation task is not an online task).
\end{itemize}

\subsection{Limitations}
This paper claims two limitations of the proposed approach: (1) CenterLineDet is restricted by the perspective transformation performance. CenterLineDet consists of two stages and cannot be trained in an end-to-end manner, which may degrade the network performance. If the perspective transformation module presents inferior BEV heatmaps, CenterLineDet would be negatively affected; (2) Although CenterLineDet presents the best performance in the experiments, it still cannot handle too complicated intersection areas very well. We aim to solve this problem by applying more powerful perspective transformation models.

\section{Conclusions and Future Work}
\label{sec:conclusion}
We presented here CenterLineDet to automatically generate lane centerline HD maps using vehicle-mounted sensors. The key problem is to detect the lane centerline graph with complicated topology. Taken as input data sequences from multiple sensors, CenterLineDet first predicts the BEV segmentation heatmap of lane centerlines. Then, a decision-making transformer network is trained to control an agent to explore the scene to create the lane centerline graph vertex by vertex. After processing all frames in the input data sequence, the trajectory of the agent was outputted as the lane centerline graph to generate HD map. The effectiveness and superiority of CenterLineDet were demonstrated by the comparative experiments on the nuScenes dataset. In the future, we plan to adopt more powerful perspective transformation models and make CenterLineDet end-to-end trainable. 

\section{Acknowledgement}
\label{sec:acknowledgement}
This work was supported in part by the Open Research Project of Zhejiang Lab under grant 2021NL0AB01, and in part by the Guangdong Basic and Applied Basic Research Foundation under Grant 2022A1515010116.

\bibliographystyle{IEEEtran}
\bibliography{mybib}
\end{document}